\documentclass[final]{svjour3}

\usepackage{times}
\usepackage{xcolor}
\usepackage{soul}
\usepackage[utf8]{inputenc}

\usepackage{siunitx}
\usepackage{booktabs}
\usepackage{csvsimple}
\usepackage{multirow}
\usepackage{graphicx}
\usepackage[caption=false]{subfig}

\usepackage[hyphens, spaces, obeyspaces]{url}

\usepackage{color, colortbl}
\usepackage{amsmath}
\usepackage{xcolor}

\definecolor{Gray}{gray}{0.9}

\definecolor{Myyellow}{RGB}{255,255, 189}
\definecolor{Myblue}{RGB}{193,240, 255}

\title{DeepFall -- Non-invasive Fall Detection with Deep Spatio-Temporal Convolutional Autoencoders}

\author{Jacob Nogas,  Shehroz S. Khan, Alex Mihailidis}
\institute{Jacob Nogas 
    \at University of Toronto, Canada \\\email{jacob.nogas@mail.utoronto.ca}
        \and Alex Mihailidis
    \at University of Toronto, Canada \\\email{alex.mihailidis@utoronto.ca}
    \and 
    Shehroz S. Khan 
    \at Toronto Rehabilitation Institute, University Health Network, Canada 
    \\\email{shehroz.khan@uhn.ca}
    }

\begin{document}

\maketitle

\begin{abstract}
  Human falls rarely occur; however, detecting falls is very important from the health and safety perspective. Due to the rarity of falls, it is difficult to employ supervised classification techniques to detect them. Moreover, in these highly skewed situations it is also difficult to extract domain specific features to identify falls. In this paper, we present a  novel framework, \textit{DeepFall}, which formulates the fall detection problem as an anomaly detection problem. The \textit{DeepFall} framework presents the novel use of deep spatio-temporal convolutional autoencoders to learn spatial and temporal features from normal activities using non-invasive sensing modalities. We also present a new anomaly scoring method that combines the reconstruction score of frames across a temporal window to detect unseen falls. We tested the \textit{DeepFall} framework on three publicly available datasets collected through non-invasive sensing modalities, thermal camera and depth cameras and show superior results in comparison to traditional autoencoder methods to identify unseen falls.
  
 
\end{abstract}

\section{Introduction}

Each year, one out of five falls that older adults incur, causes a serious injury to them such as broken bones or a head injury \cite{cdc}. Among nursing home residents, on an average, $2.6$ falls per person per year occur \cite{rubenstein1990value}. 
Detecting falls is important from the perspective of health and safety; however, due to the infrequent occurrence of falls, it is difficult to collect sufficient training data for them \cite{khan2017detecting}. Given this lack of training data, there may be no or little training data for falls, and it will be hard to employ supervised classification techniques. 
There are also concerns that the sensing devices for the task of fall detection may be invasive and breach the privacy of a person \cite{mercuri2016healthcare,yusif2016older}. Some non-invasive sensing devices, such as a thermal or depth camera, may not fully reveal the identity of a person, but it is difficult to extract discriminative features to identify unseen falls, especially in a highly skewed data scenario \cite{skubic2016testing}. 

The research work discussed in this paper deals with identifying falls in video sequences captured from non-invasive sensing modalities, such as thermal and depth cameras, which may fully or partially protect the privacy of a person.
We propose to address the problem of fall detection as an anomaly detection problem by considering the lack of fall data and the abundance of normal activities of daily living (ADL). The general idea explored in the paper is to train an autoencoder on normal activities and use their reconstruction error to identify unseen falls during testing.
A deep autoencoder (DAE) can be used to learn features from normal ADL; however, it ignores the 2D structure of  images, and may force each feature to be global to span the entire visual field \cite{masci2011stacked}. In visual recognition tasks, convolutional autoencoders (CAE) perform better because they can discover localized spatial features that repeat themselves over the input \cite{masci2011stacked}. A video sequence embeds information in both space and time; therefore, spatio-temporal convolutional autoencoders are more relevant as they can learn a representation of local spatio-temporal patterns of frames in a video \cite{baccouche2012spatio}. 

In this paper, we present a novel fall detection framework, \textit{DeepFall}, which comprises of (i) formulating fall detection as an anomaly detection problem, (ii) designing a deep spatio-temporal convolutional autoencoder (DSTCAE) and training it on \textit{only} the normal ADL, and (iii) proposing a new anomaly score to detect unseen falls. The DSTCAE first encodes stacks of contiguous frames in a time window by performing 3D convolution/3D pooling. During the decoding phase, the DSTCAE uses 3D UpSampling, or 3D deconvolution to reconstruct the input window of frames. Reconstruction error for each frame within a window is then calculated. We then present a new method to compute anomaly score, termed as within-context score, which considers the reconstruction error of frames within a window and gives an anomaly score of a given window. The anomaly score can be used to identify an unseen fall during the testing phase of DSTCAE. To the best of our knowledge, this is the first time a DSTCAE has been used for fall detection problem. 
The present paper is an extension of our previous work \cite{nogasfall2018}, and differs from it in the following manner:

\begin{itemize}
    \item In the previous work, we employed Convolutional LSTM  based Autoencoder (CLSTMAE). By contrast, in this paper we use a different 3D Convolutional autoencoder, DSTCAE, for learning spatio-temporal features from the normal ADL. Our results show that DSTCAE gives better results than CLSTMAE for all the datasets. 
    \item The previous paper only tested CLSTMAE on thermal dataset, whereas in this paper we extend our experiments on two more depth camera based fall detection datasets. 
    \item The previous paper presented a frame based anomaly score. However, in this paper we present a new anomaly scoring method that takes the decision at the level of a window (comprising of contiguous frames) and not individual frames, which resulted in similar or higher area under the curve (AUC) across all the datasets.
    \item In this paper, we did an additional experiment to understand the impact on AUC of designating an entire window as a fall by increasing the number of fall frames in it from $1$ to the maximum size of the window. This experiment provided insights on choosing appropriate fall frames in a window before identifying it as a fall window using a given anomaly score. 
    \item In the present paper, we compare CLSTMAE with three variants of DSTCAE for all the three datasets, and also added new results from one-class nearest neighbours on the features learned through the DAE.

\end{itemize}


\section{Related Work}
\label{sec:related}

To the best of our knowledge, fall detection has not been addressed as an anomaly detection problem using unsupervised deep learning methods. Therefore, in this section, we present a review of literature on video anomaly detection, which is closer in concept to the work presented in this paper.

Ribeiro et al. \cite{ribeiro2017study} propose to use CAE for detecting anomalies in videos. They extract appearance and motion features from video frames, and combine them to present different scenarios to the CAE. They train the CAE on only normal frames and use a regularized reconstruction error as a score to identify normal and anomalous frames. They extend their work by using reconstruction error from the CAE trained on normal video frames as an input to a one-class SVM and showed similar results \cite{gutoski2017detection}. Tran and Hogg \cite{tran2017anomaly} use CAE and one-class SVM for video anomaly detection. They first extract foreground optical flow patches using an empirical threshold, then train a convolutional autoencoder on these patches, which learn hierarchical sparse representations in an unsupervised manner. After training the model, sparse features are extracted and fed to a one-class SVM to learn the concept of normality. Their method shows competitive performance on two datasets in comparison to existing state-of-the-art methods. Chalapathy et al. \cite{chalapathy2017robust} present a robust CAE that considers anomalies in the data and modify the objective function for training the model. They use the reconstruction error as a score and comment that their method is not over-sensitive to anomalies; can discover subtle anomalies and can be potentially deployed in live settings. They show their results on a video activity detection dataset and other image datasets. Hasan et al. \cite{hasan2016learning} use both hand-crafted spatio-temporal features and fully convolutional feed-forward autoencoder to learn regular motion patterns in videos. They introduce a regularity score that scales the reconstruction error of a frame between $0$ and $1$. They show competitive performance of their method to other state-of-the-art anomaly detection methods. Munawar et al. \cite{munawar2017spatio} present a method  to generate unbiased features by unsupervised learning for finding anomalies in industrial robots surveillance task. They cluster the input image data based on different image cues (such as color and gradient) to generate pseudo-class labels. Then, they simultaneously train the network on different pseudo-class labels to learn neutral feature representation using convolution neural networks. Finally, they use a deep long short term memory based recurrent neural network to predict the next video frame in the learned feature space. If it deviates significantly from the observed frame, it is identified as an anomaly in time and space. A fusion of appearance as well as motion features is used by Xu et. al. \cite{xu2017detecting} to detect anomalies in video. More precisely, stacked denoising DAE networks are applied separately to optical flow, and raw image patches. Anomalies in video are then detected in a one-class SVM framework. 

Zhao et al. \cite{Zhao2017Spatio} present a spatio-temporal autoencoder that uses 3D convolutions to extract spatio-temporal features from videos. They further introduce a weight-decreasing prediction loss for generating future frames to improve the learning of motion features in videos. They tested their method on a dataset comprising of a set of real-world traffic surveillance videos along with other standard anomaly detection datasets and show superior performance on state-of-the-art approaches. Penttil{\"a} \cite{penttila2017method} propose to use 3D CAE to learn spatial and temporal features to find anomalies in hyperspectral data. After training the network, they extract feature maps from the data and feed to standard anomaly detection algorithms. A two stage cascade of autoencoders is used by Sabokrou et al. \cite{sabokrou2017deep} to detect anomalies in video. In particular, spatio-temporal patches are fed to a 3D autoencoder, where regions of interest are detected. The proposed regions are then fed to a 3D convolutional neural network. The layers of the cascaded deep networks are designed as single-class Gaussian classifiers, which are used to detect anomalies in the video.

Chong and Tay \cite{chong2017abnormal} present a method to detect anomalies in videos that consists of a spatial feature extractor and a temporal encoder-decoder framework. The spatial feature extractor comprises of convolutional and deconvolutional layers, whereas the temporal encoder-decoder is a three-layer convolutional long short term memory model. This model will be referred to as CLSTMAE. Their model is trained only on the videos with normal scenes. They use a regularity score \cite{hasan2016learning} to identify anomalies. They show comparable performance in comparison to other standard methods; however, it may produce more false alarms. 

The literature review suggests that spatio-temporal CAE are a good candidate to learn spatial and temporal features from normal activities and identify anomalies in videos. 
We expand on the previous works in the following ways
\begin{itemize}

\item None of the reviewed methods are used for fall detection problem. In \textit{DeepFall}, we train a DSTCAE only on normal ADL, and use a new anomaly scoring methods to identify a fall as anomaly from non-invasive sensors with superior performance

\item In the above related works which use reconstruction error in videos (and thus regularity score), an anomaly score is given strictly on a per frame basis. That is, a video sequence is reconstructed, and each reconstructed frame is used an an anomaly score for that frame. It is not obvious that using this score is ideal for the fall detection problem. We thus explore alternatives through our novel anomaly scoring scheme, which allows aggregating anomaly scores across overlapping sliding windows. In this scheme, we can find an anomaly score per frame, and that incorporates information from adjacent windows. Further, in this scheme we also find an anomaly score on a per-window basis. 

\item The above authors use a regularity scoring system \cite{hasan2016learning} as an anomaly score. The regularity score is a normalized reconstruction error. This normalization requires knowledge of the maximum and minimum reconstruction error for a video. Obtaining these maximum and minimum reconstruction error values requires reconstruction error for an entire video. This means that to give an anomaly score for a frame, we require reconstruction error values for all future frames of the video which this frame belongs to. Finding these future reconstruction error values is not practically feasible. We thus use un-normalized reconstruction error in our anomaly scoring scheme.
\end{itemize}
The reader may consult other papers on general anomaly detection that may involve methods besides neural networks \cite{Chandola:2009:ADS:1541880.1541882}, \cite{DBLP:journals/corr/KhanM13}, \cite{Goldstein2016ACE}

In the next section, we introduce the different components of the \textit{DeepFall} framework, which consists of a DSTCAE and anomaly scoring methods (after formulating fall detection in an anomaly detection setting).

\section{Deep Spatio-Temporal Convolutional Autoencoders}
\label{sec:stcae}

Traditional autoencoders consist of fully connected layers \cite{khan2017detectinga}. When passing an image to such a network, it is flattened into a 1D vector of pixels, and each pixel is connected to the units of the next layer. This results in a large number of network parameters, and learns spatially global features \cite{masci2011stacked}. A CAE exploits the 2D spatial structure of an image, by sharing a set of weights across the image, resulting in far fewer parameters than a traditional autoencoder, and allowing the extraction of local features from neighbouring pixels \cite{masci2011stacked}. A CAE may be a good choice to find spatial structures in images, but it cannot capture temporal information present in the contiguous frames of a video \cite{Zhao2017Spatio}. 3D-convolutions can be used to extract both temporal and spatial features by convolving a 3D kernel with the cube formed by stacking temporally contiguous frames of a video (we refer it as a window). This allows information across these contiguous frames to be connected to form feature maps, thereby capturing spatio-temporal information encoded in these adjacent frames \cite{ji20133d}. This idea can be used to construct a 
DSTCAE, where the encoding phase consists of dimensionality reduction both in space and time, of a window of images that are joined contiguously. In the decoding phase, the network attempts to reconstruct the window. 

\subsection{Sliding Window Reconstruction}
\label{sec:sw}
The inputs to DSTCAE are windows of contiguous video frames. These windows are generated by applying a temporal sliding window to video frames, with window length $T$, padding (or not) and stride $B$ (which represents the amount of frames shifted from one window to the next). If a video contains $V$ frames, and padding is not used, then the number of windows ($D$) generated is

\begin{equation}\label{eq:1}
\begin{aligned}
 D = {\lfloor\frac{V-T}{B}\rfloor+1}
 \end{aligned}
\end{equation}

In our implementation of DSTCAE, we chose  $T = 8$ (this choice is explained in Section \ref{sec:processing}),  no padding and $B = 1$, which ensures that all the frames of the video are selected in forming windows. We re-size all the input frames to $64 \times 64$. The resulting network input ($I$) thus have dimensions $8 \times 64 \times 64$. DSTCAE encodes $I$ into spatio-temporal features, and then attempts to reconstruct it from the encoded representation. For training DSTCAE, we use mean squared error loss between $I$ and reconstructed output window $O$ (same dimensions as $I$), optimized on a per batch basis, giving the following cost function:

\begin{equation}\label{eq:2}
     C(\theta) = \frac{1}{N}\sum_{i=1}^{N}\Vert I_i-O_i \Vert_2^2
\end{equation}

where $N$ is the number of training samples in a batch, $\theta$ denotes the network parameters, and $\Vert \cdot \Vert_2$ denotes the Euclidean norm.



\subsection{3D Convolutions}
The main component of a DSTCAE is the 3D convolutional layer, which is defined as follows: the value $v$ at position $(x,y,z)$ of the $j^{th}$ feature map in the $i^{th}$ 3D convolution layer, with bias $b_{ij}$, is given by the equation \cite{ji20133d}

\begin{equation}\label{eq:3}
    v_{ij}^{xyz} = f\left(\sum_m \sum_{p=0}^{P_i-1} \sum_{q=0}^{Q_i-1} \sum_{s=0}^{S_i-1}  w_{ijm}^{pqs}v_{(i-1)m}^{(x+p)(y+q)(z+s)} + b_{ij}\right)
\end{equation}

where $P_i$, $Q_i$, $S_i$ are the vertical (spatial), horizontal (spatial), and temporal extent of the filter cube $w_i$ in the $i^{th}$ layer. The set of feature maps from the $(i-1)^{th}$ layer are indexed by $m$, and $w_{ijm}^{pqs}$ is the value of the filter cube at position $pqs$ connected to the $m^{th}$ feature map in the previous layer. Multiple filter cubes will output multiple feature maps. 

\subsection{3D Encoding and Decoding}
\label{sec:3DCAE-encdec}

\paragraph{Encoding}: The input $I$ introduced in Section \ref{sec:stcae} in a DSTCAE is encoded by a sequence of 3D convolution and 3D-max-pooling layers. 3D convolutions operate as described in equation \ref{eq:3}, with stride of $1 \times 1 \times 1$, and padding. The max-pooling layers use padding, with stride and kernel dimensions $2 \times 2 \times 2$. This means that each dimension (temporal depth, height, and width) is reduced by a factor of $2$ with every max-pooling layer. This process is repeated for $2$ level of depth.  
For hidden layers (in both encoding and decoding), the activation function $f$ in equation \ref{eq:3} is set to $\mathbf{ReLU}$.  We use $P_i = Q_i = 3$, and $S_i = 5$, for all convolutional and deconvolutional layers, as these values were found to produce the best results across all data sets. The specification of the encoding and decoding process is shown in Table \ref{tab:config3D} and described in detail below.

\begin{table*}[ht]
\centering
\begin{tabular}{| m{38mm} | m{38mm}|m{38mm}|}\hline
 \multicolumn{1}{|c|}{\textbf{DSTCAE-UpSampling}} &
 \multicolumn{1}{|c|}{\textbf{DSTCAE-Deconv}} &
 \multicolumn{1}{|c|}{\textbf{DSTCAE-C3D}}  \\ \hline\hline
Input - (8, 64, 64, 1) & Input - (8, 64, 64, 1) & Input - (8, 64, 64, 1) \\ \hline

\cellcolor{Myblue}3D Convolution  - (8, 64, 64, 16) & \cellcolor{Myblue}3D Convolution  - (8, 64, 64, 16) & \cellcolor{Myblue}3D Convolution  - (8, 64, 64, 16)\\ \hline
\cellcolor{Myblue}3D Max-pooling - (4, 32, 32, 16) & \cellcolor{Myblue}3D Max-pooling - (4, 32, 32, 16) & \cellcolor{Myblue}3D Max-pooling - (8, 32, 32, 16)\\ \hline
\cellcolor{Myblue}3D Convolution - (4, 32, 32, 8) & \cellcolor{Myblue}3D Convolution - (4, 32, 32, 8) & \cellcolor{Myblue}3D Convolution  - (8, 32, 32, 8)\\ \hline
\cellcolor{Myblue}3D Max-pooling - (2, 16, 16, 8) & \cellcolor{Myblue}3D Max-pooling - (2, 16, 16, 8) & \cellcolor{Myblue}3D Max-pooling - (4, 16, 16, 8)\\ \hline
\cellcolor{Myyellow}3D Convolution - (2, 16, 16, 8) & \cellcolor{Myyellow}3D Deconvolution - (4, 32, 32, 8) & \cellcolor{Myblue}3D Convolution  - (4, 16, 16, 8) \\ \hline
\cellcolor{Myyellow}3D UpSampling - (4, 32, 32, 8) &  \cellcolor{Myyellow}3D Deconvolution - (8, 64, 64, 16) & \cellcolor{Myblue}3D Max-pooling - (2, 8, 8, 8)\\ \hline
\cellcolor{Myyellow}3D Convolution - (4, 32, 32, 8) & \cellcolor{Myyellow}3D Deconvolution - (8, 64, 64, 1)   & \cellcolor{Myyellow}3D Convolution - (2, 8, 8, 8)\\ \hline
\cellcolor{Myyellow}3D UpSampling - (8, 64, 64, 16)& & \cellcolor{Myyellow}3D UpSampling  - (4, 16, 16, 8)\\ \hline
\cellcolor{Myyellow}3D Convolution - (8, 64, 64, 1)& & \cellcolor{Myyellow}3D Convolution  - (4, 16, 16, 8)\\ \hline
& & \cellcolor{Myyellow}3D UpSampling  - (8, 32, 32, 8)\\ \hline
& & \cellcolor{Myyellow}3D Convolution  - (8, 32, 32, 8)\\ \hline
& & \cellcolor{Myyellow}3D UpSampling  - (8, 64, 64, 16) \\ \hline
& & \cellcolor{Myyellow}3D Convolution  - (8, 64, 64, 1) \\ \hline

\hline

\end{tabular}
\caption{Configuration of the encoding and decoding of DSTCAE-UpSampling, DSTCAE-Deconv, and DSTCAE-C3D. Encoding is in blue, and decoding in yellow}
\label{tab:config3D}
\end{table*}

\begin{figure}[ht!]
     \centering
     \includegraphics[width=90mm,height=60mm, keepaspectratio=true]{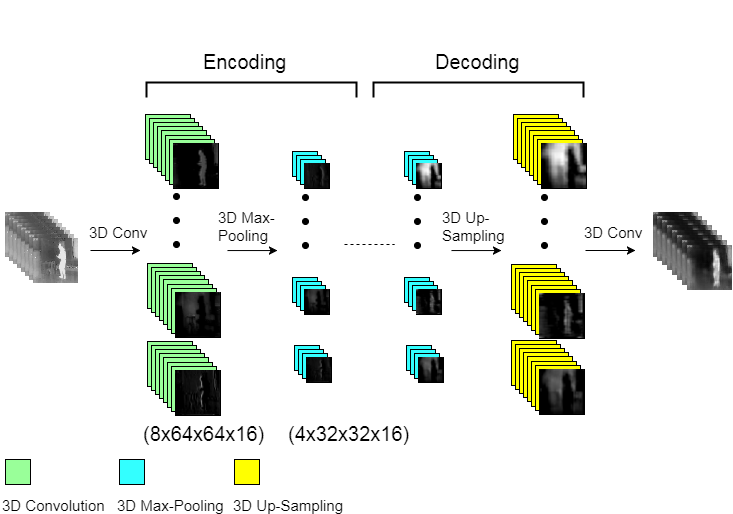}
     \caption{Outline of DSTCAE layers using fixed UpSampling (DSTCAE-UpSampling). Each layer has dimensions (temporal depth $\times$ height $\times$ width $\times$ number of feature maps)}
     \label{fig:3DCAE}
 \end{figure}

\paragraph{Decoding}: For the decoding of DSTCAE, we explore two variants. 
The first method (DSTCAE-UpSampling) uses padded 3D convolutions with stride $2 \times 2 \times 2$, followed by a fixed UpSampling operation for increasing dimensions (as seen in Figure \ref{fig:3DCAE}). In particular, we use 3D UpSampling layers as defined in Keras \cite{chollet2015keras}, with UpSampling factors $2 \times 2 \times 2$. That is, matrix elements are repeated across each dimension, such that the extent of all dimensions is doubled. The second method (DSTCAE-Deconv) uses 3D deconvolutions \cite{chollet2015keras}, with stride $2 \times 2 \times 2$, and padding \cite{Zhao2017Spatio} instead of UpSampling. This results in an increase in each dimension by a factor of $2$, thus undoing a max-pooling operation. For both the decoding methods, the output layer activation function $f$ is set to $\mathbf{tanh}$; computed element-wise in both cases. We use $\mathbf{tanh}$ for the output layer to limit the reconstructed pixel values in the range $[-1,1]$, so that they are comparable to the input 
(see Section \ref{sec:processing}). In both decoding methods, outputs of the network ($O$) have the same dimensions as $I$, and are generated by convolving a single filter cube with a final layer of feature maps. 
By proposing two methods for the decoding procedure in DSTCAE, we created two variants of the spatio-temporal autoencoder, whose performance are compared for detecting unseen falls in Section \ref{sec:results}. 

 A diagram illustrating the DSTCAE-UpSampling structure is shown in Figure \ref{fig:3DCAE}. The first layer generates $16$ feature maps (depicted in figure by vertical stacks), by convolving $16$ different filter cubes with the input window of frames. This layer is followed by a 3D max-pooling layer. Convolution and pooling layers are repeated in this manner once more, resulting in an encoded dimension of  $2 \times 16 \times 16 \times 8$. We choose to stop encoding at this point, in order to avoid collapsing the temporal dimension completely; a larger value of $T$ would allow for a deeper network. Decoding then involves two UpSampling/3D convolution steps, bringing the hidden representation dimensions to $8 \times 64 \times 64 \times 16$. A final 3D convolution layer combines these $16$ feature maps into the decoded reconstruction $O$.

We also test a third 3D convolutional autoencoder variant, based on Tran et al. \cite{C3D}, which we refer to as DSTCAE-C3D. The DSTCAE-C3D network has the same encoding and decoding as DSTCAE-UpSampling, but with an extra 3D Convolution/3D Max-pooling layer in encoding (see table \ref{tab:config3D}), and extra 3D UpSampling/3D convolution in decoding. The extra 3D Max-pooling has padding, with stride and kernel dimensions $1 \times 2 \times 2$. This results in spatial dimension reduction, but not temporal; allowing for greater network depth without collapsing the temporal dimension.

To summarize, we compare three variants of DSTCAE: 1) DSTCAE-UpSampling: uses UpSampling for decoding, 2) DSTCAE-Deconv: uses deconvolution for decoding (similar to \cite{Zhao2017Spatio}), and  3) DSTCAE-C3D: uses UpSampling for decoding, but contains additional spatial pooling/unpooling layer. These architectures are summarized in Table \ref{tab:config3D}. In this table, encoding is highlighted in blue, and decoding in yellow.

We compare these DSTCAE variants to 2D frame based models CAE and DAE. Their specifications are shown in Table \ref{tab:config}. This table only shows the configuration of the encoding phase of the autoencoder. The decoding configuration is the same, but using UpSampling, or deconvolution for CAE (referred to as CAE-UpSampling, and CAE-Decconv respectively), and more fully connected layers for DAE. Also, dropout is applied to layer $1$ for DAE. We also use dropout for layer $2$ for all DSTCAE variants. In all cases the dropout probability is set to $0.25$. Dropout was not found to be beneficial for CAE models, and so was not used.

\begin{table}[ht]
\centering
\begin{tabular}{| m{42mm} | m{35mm}|}\hline
 \multicolumn{1}{|c|}{\textbf{CAE}} & \multicolumn{1}{|c|}{\textbf{DAE}}  \\ \hline\hline
Input - (64, 64, 1)  &  Input - (64,64,1)   \\ \hline
2D Convolution - (64, 64, 16) & Fully Connected - (4096)  \\ \hline
2D Max-pooling - (32, 32, 16) & Fully Connected - (150) \\ \hline
2D Convolution - (32, 32, 8) & Fully Connected - (100) \\ \hline
2D Max-pooling - (16, 16, 8) & Fully Connected - (50) \\ \hline
2D Convolution - (16, 16, 8) & - \\ \hline
2D Max-pooling - (8, 8, 8) & - \\ \hline

\end{tabular}
\caption{Configuration of the encoding phase of CAE and DAE.}
\label{tab:config}
\end{table}

\section{Anomaly Scores to Detect Unseen Falls}
\label{sec:anomaly}
In this paper, we formulated fall detection in a one-class classification framework \cite{khan2017detectinga}, where normal ADL are available in abundance and no fall data is present during training. However, during testing, both normal ADL and falls may be present. Therefore, we detect falls as an anomaly. In the \textit{DeepFall} framework, the general process for detecting unseen falls is:
\begin{itemize}
    \item Train a given type of autoencoder by minimizing reconstruction error on ADL,
    \item During testing, get reconstitution error of a frame (or a window comprising of different frames); if the reconstruction error is `high', then a fall is detected.
\end{itemize}

Previous anomaly scoring schemes involve using a regularity score \cite{hasan2016learning}, which is a normalized reconstruction error requiring future values from test frames, which is a major drawback in real world deployment. We thus strictly use un-normalized reconstruction error.

Further, previous anomaly scoring schemes \cite{chong2017abnormal} \cite{hasan2016learning}, \cite{Zhao2017Spatio} etc. compute an anomaly score on a per frame basis, using the reconstruction error of that frame. Such a scoring scheme may not be ideal for the fall detection problem, and so we propose to explore a new anomaly scoring scheme, presented in this section. In our novel anomaly scoring scheme, we compute an anomaly score per frame, while also incorporating information about reconstruction error in adjacent overlapping sliding windows. We also explore computing an anomaly score on a per-window basis. The details of this anomaly scoring scheme are as follows:

For CAE and DAE, the input to the network is a single frame; therefore, the reconstruction error is computed per frame. For all the variants of DSTCAE, the input to the network is a temporal window of $T$ video frames. That is, given a test video sequence, we apply a sliding window as described in Section \ref{sec:stcae}. For the $i^{th}$ window $I_{i}$, the network outputs a reconstruction of this window, $O_{i}$. The reconstruction error ($R_{i,j}$) between the $j^{th}$ frame of $I_i$ and $O_i$ can be calculated as

\begin{equation}
    R_{i,j}= \Vert I_{i,j} - O_{i,j} \Vert_2^2
    \label{eq:reconerror}
\end{equation}

\begin{figure}[ht!]
    \centering
    \includegraphics[scale=0.45]{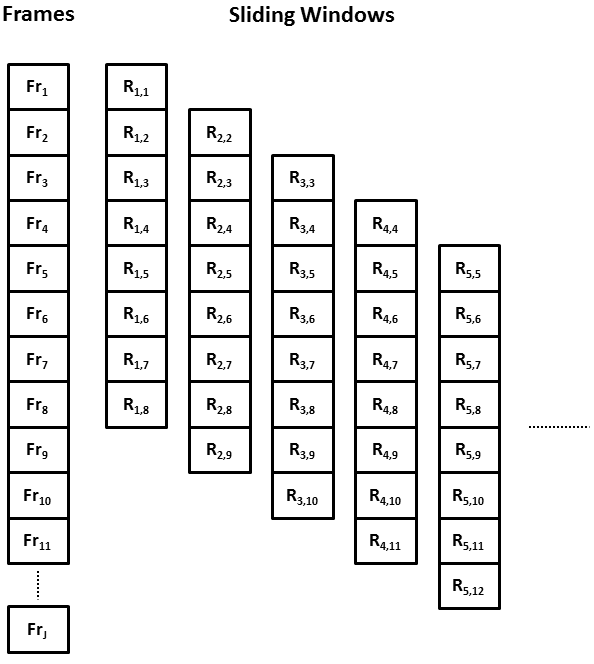}
    \caption{Temporal sliding window showing reconstruction error ($R_{i,j}$) per frame ($Fr_j$) with $T=8$.}
    \label{fig:windows}
\end{figure}

Figure \ref{fig:windows} shows the sliding window approach for $T = 8$. The first window of $T = 8$ frames, $I_{1}$ ($Fr_1$ to $Fr_8$) are reconstructed, and their corresponding reconstruction error is stored ($R_{1,1:8}$). For the next window of frames, the input window is shifted forward in time by one frame. This process continues until all frames are used.

\subsection{Cross-Context Anomaly Score}

To compare DSTCAE models with CAE and DAE, we must get a score per frame. A frame can appear in multiple windows. For instance, frame $Fr_2$ (see Figure \ref{fig:windows}) appears in two windows and, thus gives two reconstruction errors: $R_{1,2}$, and $R_{2,2}$. The former is attained when frame $Fr_2$ was the second frame in the window, and the latter case when frame $Fr_2$ was the first frame of the input window. Each window that a frame appears in provides a different temporal \textit{context} within which this frame can be viewed. 
The cross-context anomaly score gives scores on a per-frame basis, by considering all of the reconstruction errors obtained for a frame across different windows (temporal contexts) \cite{nogasfall2018}. 
For a frame $j$, an anomaly score can be calculated 
based on the mean ($C_{\mu}^{j}$) or standard deviation ($C_{\sigma}^{j}$) of the reconstruction errors ($R_{i,j}$) across different windows (or contexts):

\begin{equation}
\begin{aligned}
C_{\mu}^{j} &= \begin{cases}\frac{1}{j} \sum_{i=1}^j R_{i,j}  & j<T\\
\frac{1}{T} \sum_{i=1}^T R_{i,j} &  j \geq T
\end{cases}\\
 C_{\sigma}^{j} &= \begin{cases}\sqrt{\frac{1}{j} \sum_{i=1}^j (R_{i,j}-C_{\mu}^{j})}& j<T\\
 \sqrt{\frac{1}{T} \sum_{i=1}^T (R_{i,j}-C_{\mu}^{j})} & j \geq T
 \end{cases}
 \end{aligned}
\end{equation}

$C_{\mu}^{j}$ and $C_{\sigma}^{j}$ give an anomaly score per-frame, while incorporating information from the past and future. A large value of $C_{\mu}^{j}$ or $C_{\sigma}^{j}$ means that the $j^{th}$ frame, when appearing at different positions in subsequent windows, is reconstructed with a high average error or highly variable error. In a normal ADL case, the reconstruction error of a frame should not vary a lot with its position in subsequent windows; however, if it does, then this may indicate anomalous behaviour, such as a fall.

\subsection{Within-Context Anomaly Score}
In a case when videos are considered as sequences of contiguous frames, we may want to designate the class of the whole window as fall or normal ADL, instead of deciding the class of every frame across different windows (or contexts). This way, all the frames present in a window will be used for predicting its class and they will have no influence on subsequent windows. Therefore, we call this method as within-context anomaly score.
This method considers the reconstruction of frames within a context (or windows shown vertically in Figure \ref{fig:windows}), and gives a single score. In particular, the mean ($W_{\mu}^{i}$) and standard deviation ($W_{\sigma}^{i}$) of the reconstruction error of all the frames within a window $I_i$, can be computed as an anomaly score to identify a fall,

\begin{equation}
W_{\mu}^{i} = \frac{1}{T} \sum_{j=i}^{T+i-1} R_{i,j} ,\quad
W_{\sigma}^{i} = \sqrt{\frac{1}{T} \sum_{j=i}^{T+i-1} (R_{i,j}-W_{\mu}^{i})}
\end{equation} 

A large value of $W_{\mu}^{i}$ means that the average reconstruction error of the $T$ frames within the window is high. Similarly, a large value of $W_{\sigma}^{i}$ means that the variation of reconstruction error among the $T$ frames of a window is high. 
Both the situations indicate that high reconstruction error may be happening due to one or more frames present within a window. Therefore, these situations represent the case when DSTCAE reconstructs the $T$ frames in the window with high/variable reconstruction error, which may represent an anomalous behaviour such as a fall. 

In order to designate a whole window of frames as a fall or non fall, we set a threshold on the number of fall frames in the window. That is, if the sequence contains more fall frames than the threshold, the sequence is labelled as a fall. Otherwise it is labelled as a non-fall window. If this threshold is too low then \textit{DeepFall} may be very sensitive, and may generate many false positives. Whereas, if this threshold is too high (for example all frames in the sequence are required to be fall frames), then our system may miss some falls. Therefore, this threshold, which is the number of fall frames that are required for a window of frames to be designated as a fall, is a hyper-parameter of the \textit{DeepFall} framework, and is discussed in detail in Section \ref{sec:results}.

\section{Experiments and Results}
\label{sec:experiments}

\subsection{Datasets}

We test the \textit{DeepFall} framework on the following three publicly available data sets that collected normal ADL and falls using non-invasive sensing modalities. 

\subsubsection{Thermal Fall Dataset}
\begin{figure}[!ht]
\centering
\captionsetup[subfigure]{width=80pt, justification=centering}%
  \begin{subfloat}[]
    \centering
    \includegraphics[scale=0.15]{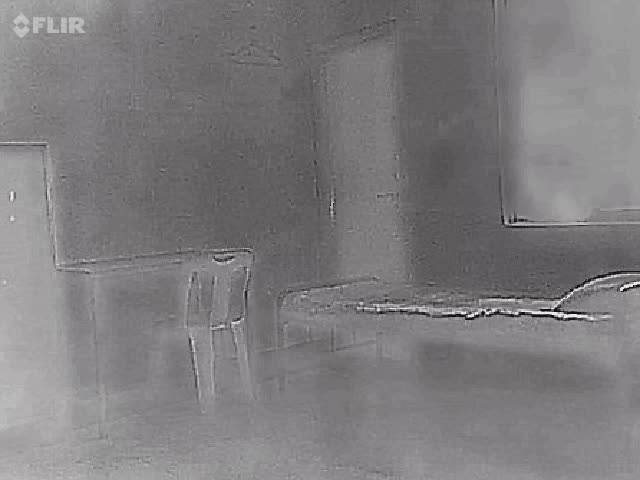}
    \label{fig:ThermalSample1}
  \end{subfloat}
  \begin{subfloat}[]
    \centering
    \includegraphics[scale=0.15]{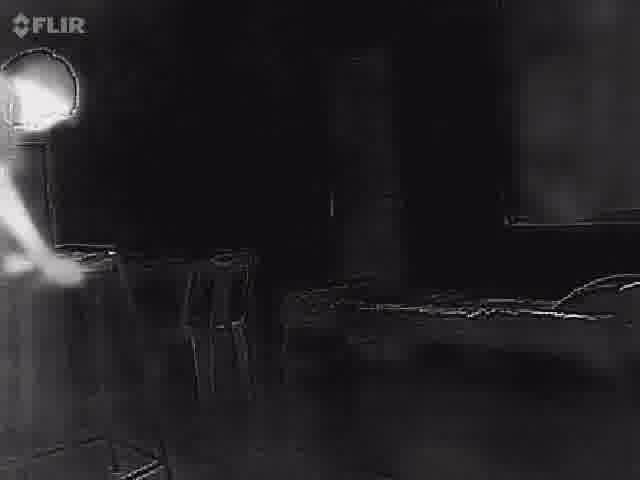}
    \label{fig:ThermalSample2}
  \end{subfloat}
    \begin{subfloat}[]
    \centering
    \includegraphics[scale=0.15]{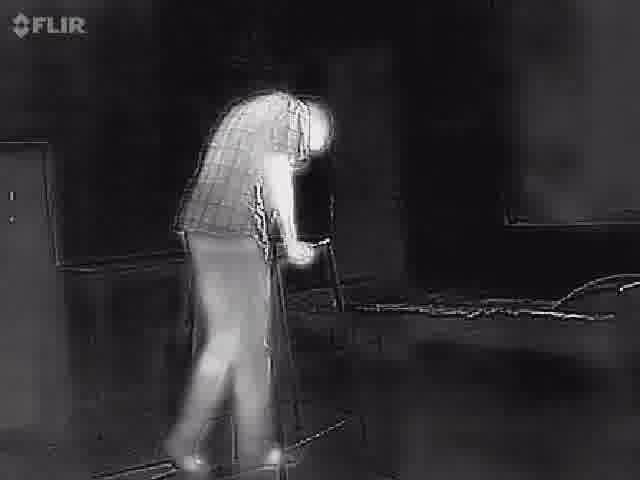}
    \label{fig:ThermalSamples3}
  \end{subfloat}
  \vspace{5mm}
      \begin{subfloat}[]
    \centering
    \includegraphics[scale=0.15]{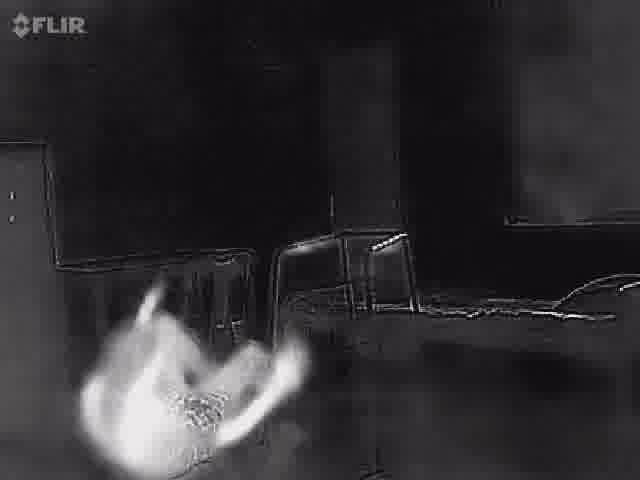}
    \label{fig:ThermalSamples4}
  \end{subfloat}
  \begin{subfloat}[]
    \centering
    \includegraphics[scale=0.15]{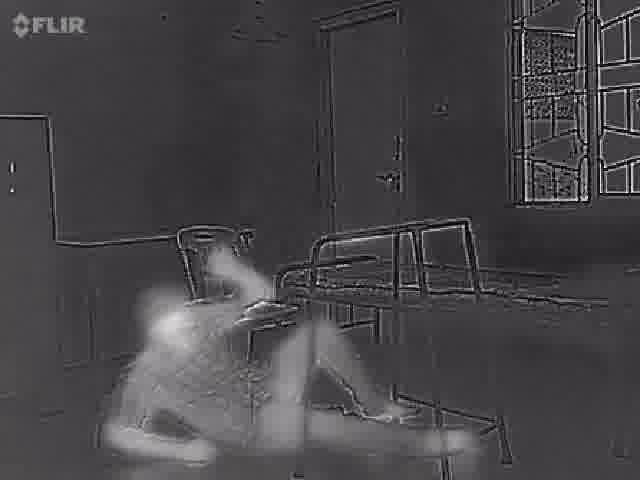}
    \label{fig:ThermalSamples5}
  \end{subfloat}
  \begin{subfloat}[]
    \centering
    \includegraphics[scale=0.15]{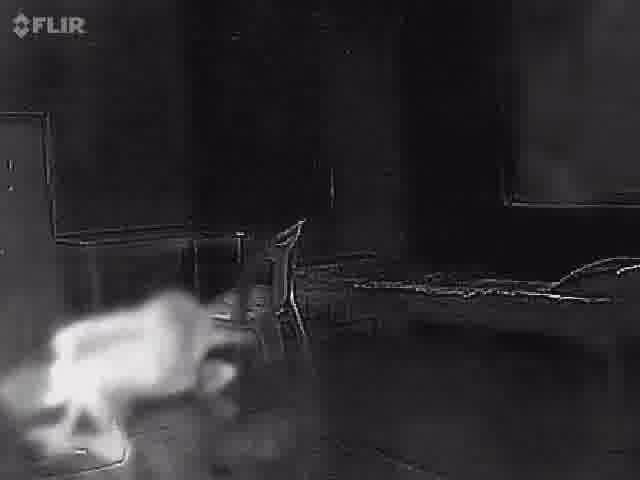}
    \label{fig:ThermalSamples6}
  \end{subfloat} 
  \centering
  \caption{Thermal Dataset -- ADL frmes (a) Empty Scene, (b) Person Entering the Scene, (c) Person in the Scene,  and Fall Frames (d), (e), and (f).}
  \label{fig:ThermalSamples}
\end{figure}

This dataset consists of videos captured by a FLIR ONE thermal camera mounted on an Android phone in a room setting with a single view \cite{Thermal}. The videos have a frame rate of either $25$ fps or $15$ fps, which was obtained by observing the properties of each video. A total of $44$ videos are collected, out of which $35$ videos contain a fall along with normal ADL, and $9$ videos contain only ADL. The spatial resolution of the thermal images is $640 \times 480$. The Thermal dataset contains many empty frames; that is, scenes where no person is present. It also contains frames of people entering the scene from the left and from the right. Examples of thermal ADL and fall frames are given in Figure \ref{fig:ThermalSamples}.

The Thermal dataset contains $22,116$ ADL frames from $9$ videos. After windowing the ADL videos individually, we generate $22,053$ windows of contiguous frames (using equation \ref{eq:1}), which are used for training spatio-temporal autoencoders.

\subsubsection{UR Fall Detection Dataset}
\begin{figure}[!ht]
\centering
\captionsetup[subfigure]{width=80pt}%
  \begin{subfloat}[]
    \centering
    \includegraphics[scale=0.1]{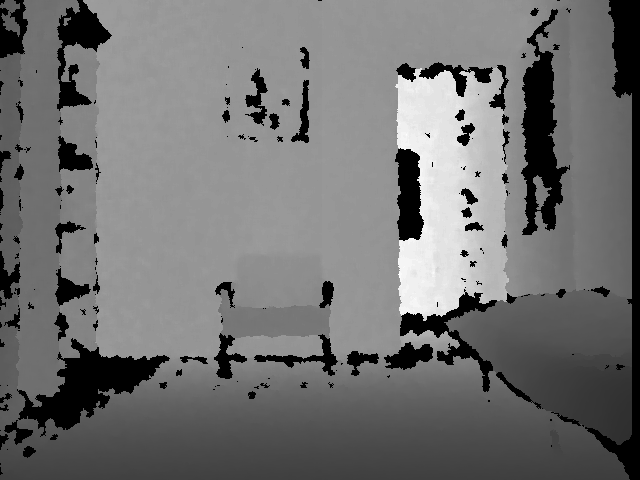}
  \end{subfloat}
    \begin{subfloat}[]
    \centering
    \includegraphics[scale=0.1]{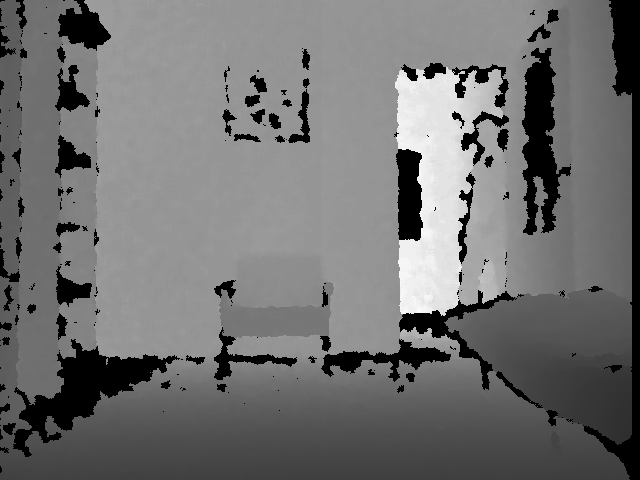}
  \end{subfloat}
  \begin{subfloat}[]
    \centering
    \includegraphics[scale=0.1]{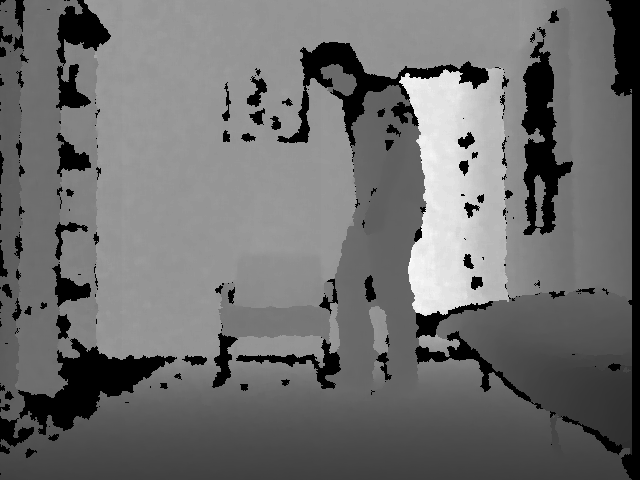}
  \end{subfloat}
  \begin{subfloat}[]
    \centering
    \includegraphics[scale=0.1]{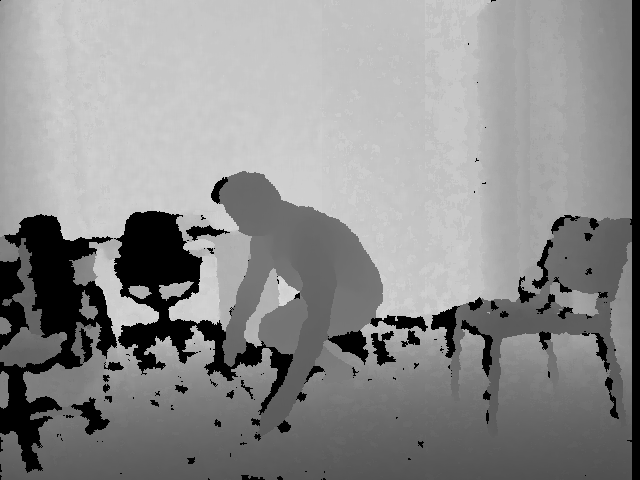}
  \end{subfloat}
  \caption{UR Dataset - Original Depth frames with holes (a) Empty Scene (b) Person entering the Scene, (c) Person in the Scene, (d) Fall.}
\label{fig:URHoles}
  \vspace{5mm}  
    \begin{subfloat}[]
    \centering
    \includegraphics[scale=0.1]{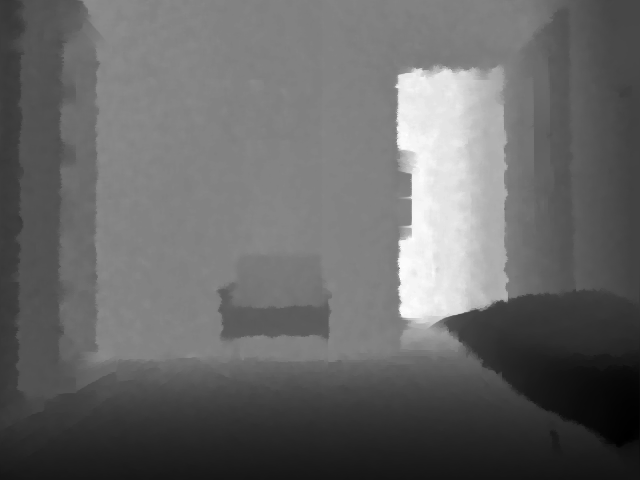}
  \end{subfloat}
    \begin{subfloat}[]
    \centering
    \includegraphics[scale=0.1]{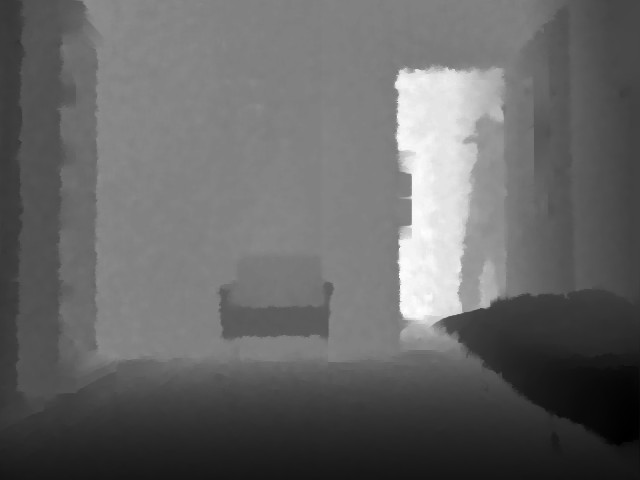}
  \end{subfloat}
    \begin{subfloat}[]
    \centering
    \includegraphics[scale=0.1]{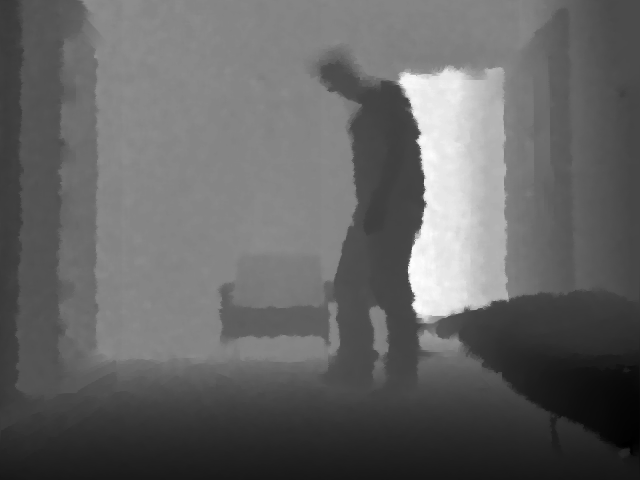}
  \end{subfloat}
  \begin{subfloat}[]
    \centering
    \includegraphics[scale=0.1]{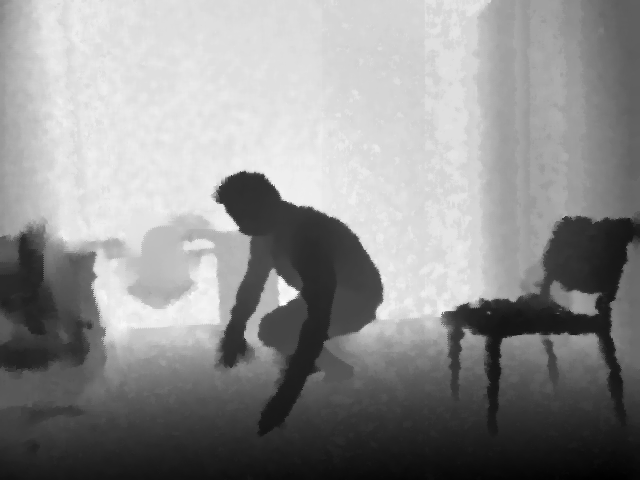}
  \end{subfloat}
    \caption{UR Dataset - Depth frames after holes filling (a) Empty Scene (b) Person entering the Scene, (c) Person in the Scene, (d) Fall.}
    \label{fig:URHolesFilled}
\end{figure}

The UR dataset (UR) \cite{UR} contains $70$ depth videos collected using a Microsoft Kinect camera at $30$ fps. that was mounted parallel to the floor. Of these, $30$ videos contain a fall, and $40$ videos contain various ADL, such as walking, sitting down, crouching down, and lying down in bed. Five persons performed two types of falls -- from standing position and from sitting on the chair. The pixels in the depth frames indicate the calibrated depth in the scene. The depth map is provided in a $640 \times 480$ resolution. The UR dataset contains empty frames. It also contains frames of people entering the scene towards the camera.
Samples of original ADL and fall depth frame of the UR dataset are shown in Figure \ref{fig:URHoles}.
After applying the sliding window, we obtain $8,661$ windows of contiguous frames used for training different spatio-temporal autoencoders.

\subsubsection{SDU Dataset}
\begin{figure}[!ht]
\centering
\captionsetup[subfigure]{width=80pt}%
  \begin{subfloat}[]
    \centering
    \includegraphics[scale=0.1]{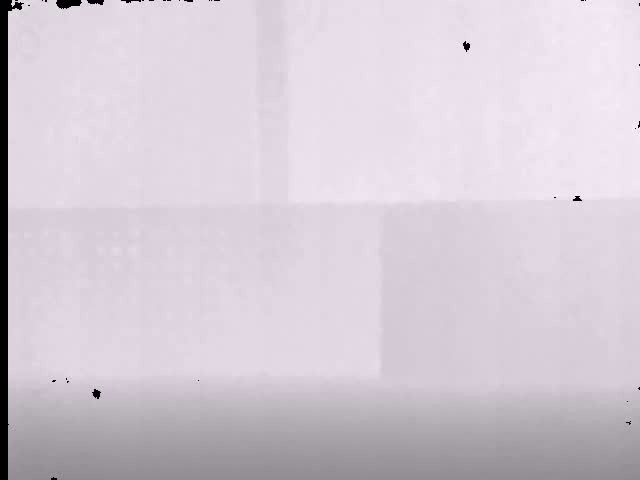}
  \end{subfloat}
 \captionsetup[subfigure]{width=80pt}%
  \begin{subfloat}[]
    \centering
    \includegraphics[scale=0.1]{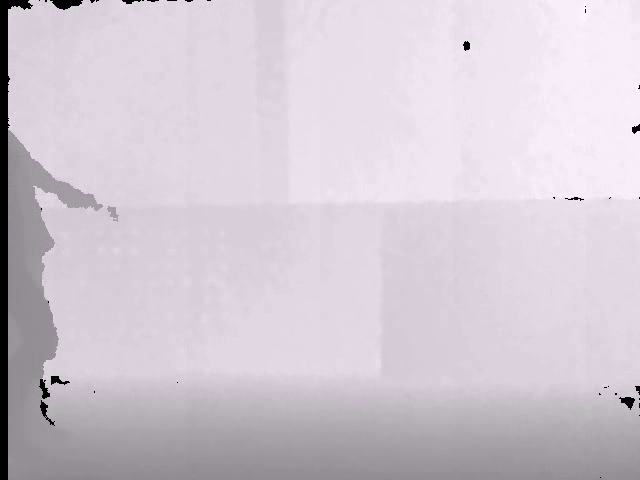}
  \end{subfloat}
  \begin{subfloat}[]
    \centering
    \includegraphics[scale=0.1]{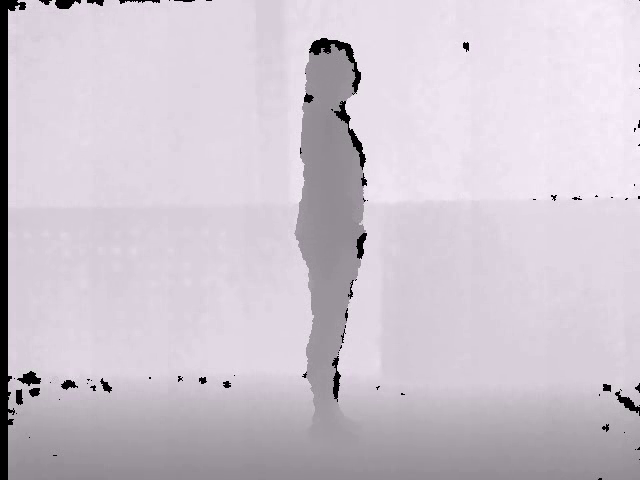}
  \end{subfloat}
   \begin{subfloat}[]
    \centering
    \includegraphics[scale=0.1]{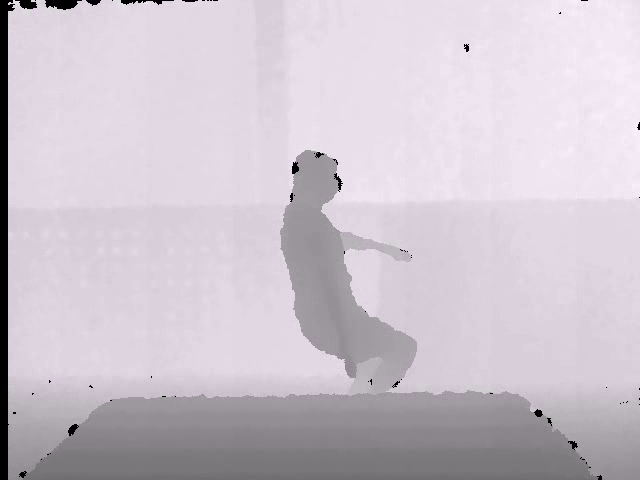}
  \end{subfloat}
  \caption{SDU Dataset - Original Depth frames with holes (a) Empty Scene (b) Person entering the Scene, (c) Person in the Scene, (d) Fall.}
  \label{fig:SDUHoles}
\vspace{5mm}  
\begin{subfloat}[]
    \centering
    \includegraphics[scale=0.1]{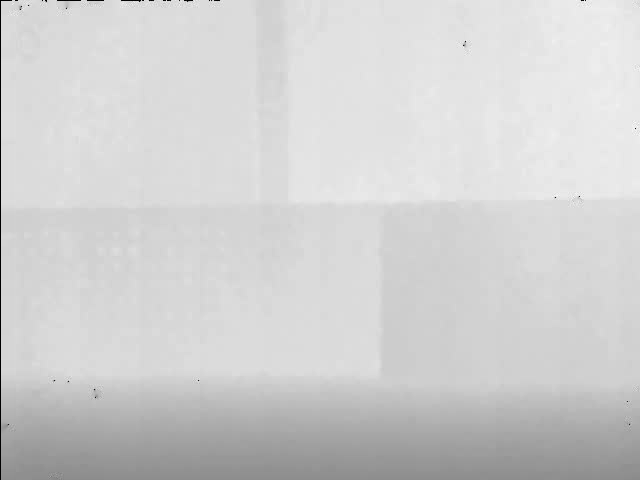}
  \end{subfloat}
 \captionsetup[subfigure]{width=80pt}%
  \begin{subfloat}[]
    \centering
    \includegraphics[scale=0.1]{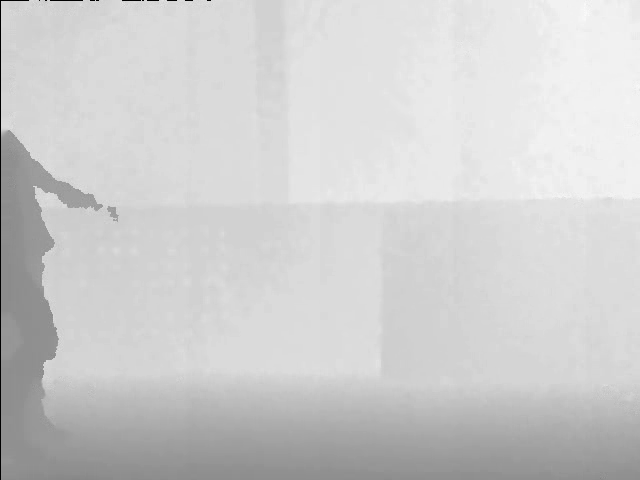}
  \end{subfloat}
  \begin{subfloat}[]
    \centering
    \includegraphics[scale=0.1]{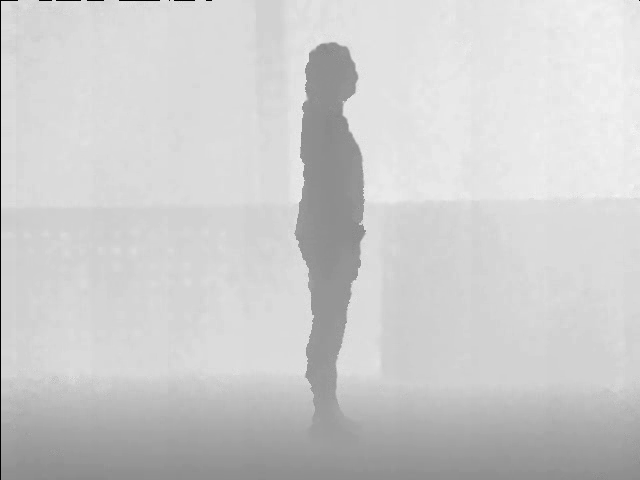}
  \end{subfloat}
  \begin{subfloat}[]
    \centering
    \includegraphics[scale=0.1]{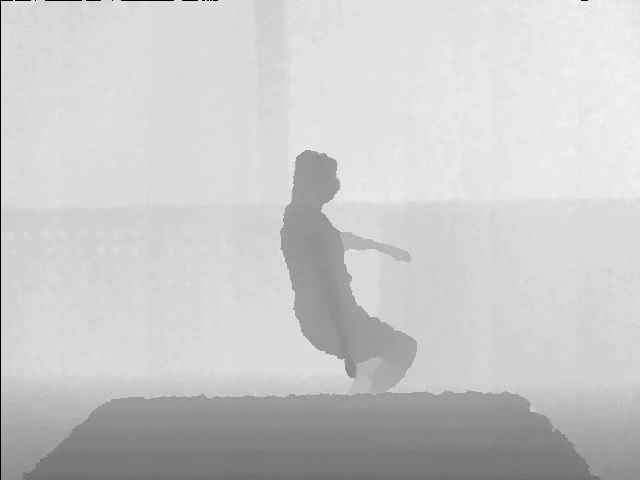}
  \end{subfloat}
  \caption{SDU Dataset - Depth frames after hole filling, (a) Empty Scene (b) Person entering the Scene, (c) Person in the Scene, (d) Fall.}
  \label{fig:SDUHolesFilled}
\end{figure}

The SDU dataset \cite{SDU}, contains depth videos collected by a Microsoft Kinect camera. The data that was shared with us contains $1,197$ depth videos. Of these videos, $997$ contain the following ADL: bending, squatting, sitting, lying, and walking. The remaining $200$ videos contain a fall, as well as other various ADL. The activities are performed by $20$ young men and women. Each person performs $10$ trials, which consist of simulating a fall, as well as performing each of the above ADL. The videos are recorded at $30$ fps, with a spatial resolution of $320 \times 240$, and an average length of $5$ seconds. After applying the sliding window, we obtain $163,573$ windows of contiguous frames used for training spatio-temporal autoencoders. The SDU dataset contains empty frames. It also contains frames of people entering the scene from the left and right. Samples of original ADL and fall depth frame from the SDU dataset are shown in Figure \ref{fig:SDUHoles}. 

\subsection{Data-preprocessing}
\label{sec:processing}
The Thermal data set frames were extracted from mp4 video files. The UR dataset's frames were already available in png format. The SDU dataset frames were extracted from AVI video files. All the frames in all the datasets are normalized by dividing the pixel values by $255$ to keep them in the range $[0, 1]$, and subtracting the per-frame mean from each frame, resulting in pixel values in the range $[-1,1]$. As discussed in Section \ref{sec:sw}, all the frames are also re-sized to $64 \times 64$. 

In order to create windows of contiguous video frames to give as input to the DSTCAE for training and testing, we perform a sliding window on all video frames, as described in \ref{sec:sw}. We choose $T = 8$ because we do not want our window length to exceed the length of the shortest fall sequence, which is found to be $13$ frames. We thus chose the largest power of $2$ smaller than $13$, which is $8$. We use the same window length for all datasets for simplicity. Smaller window length may lead to higher false positives as the decision regarding a fall would have to be made on fewer reconstructed video frames. On the other hand, a greater window length may include non-fall frames because fall is a short term event. Therefore, some falls may be missed to be detected. 

In both the UR dataset and the SDU dataset, there were many regions of missing pixels, termed as `holes'. Both of these datasets uses Microsoft Kinect depth sensor, which is a structured 3D scanner and light measurements from it can contain errors due to multiple reflections, transparent objects, occlusion or scattering \cite{viacheslav2014kinect}. These holes can be seen as black regions in Figures \ref{fig:URHoles} and \ref{fig:SDUHoles}. 
These holes can be detrimental to learning useful spatio-temporal features using the \textit{DeepFall} framework. Therefore, we employed hole filling methods to fill in these regions. 
For the UR dataset, we used a method based on depth colorization \cite{Silberman:ECCV12}. Figure \ref{fig:URHolesFilled} shows the corresponding depth frame after filling holes in the frames shown in Figure \ref{fig:URHoles}. For the SDU dataset, we did not have information of distance in the depth frames, since they were extracted from AVI formatted videos. That is, the depth images contain values in the range $[0,255]$, where darker pixels correspond to a lesser distance from the camera; true distance is not known for these values.  Distance information is required for using the depth colorization algorithm \cite{Silberman:ECCV12}. Thus, we used a simple inpainting technique, based on an OpenCV \cite{opencv} implementation of the inpainting algorithm presented in \cite{NS}. Figure \ref{fig:SDUHolesFilled} shows the results of applying this inpainting technique on the frame shown in Figure \ref{fig:SDUHoles}.

To evaluate the effectiveness of hole filling in detecting falls using different autoencoder methods, we created another version of the UR and SDU dataset in which the holes are filled for all the frames, we call them as UR-Filled and SDU-Filled.

\subsection{Experimental Setup}
Different types of autoencoders are trained on only the ADL videos, which only contain normal ADL / non-fall frames. The testing was carried out on fall videos, which contain fall activity as well as normal ADL. The video frames used for training the models were not annotated because they were all considered normal ADL. In order to test the models, fall videos are used that contained both fall and non-fall frames. The fall frames were manually annotated in these videos. 

For comparing the performance of different variants of spatial-temporal autoencoder, DAE, CAE,  all autoencoders are trained for 500 epochs. Adadelta 
 optimizer was used in training these models. The training batch size is set to $32$ for DAE and CAE models, and $16$ for DSTCAE variants, where each batch consists of a stack of $8$ frames.
To train DAE and CAE variants, we augment the data by performing horizontal flipping. No data augmentation was performed when training DSTCAE variants, as it did not improve results. As noted in Section \ref{sec:3DCAE-encdec}, dropout is applied to layer $1$ for DAE. and to layer $2$ for all DSTCAE variants. In all cases the dropout probability is set to $0.25$. \footnote{The code for DSTCAE is available at https://github.com/JJN123/Fall-Detection}.

The within-context anomaly score can only be calculated for the DSTCAE variants, because it calculates a score for each window of $T$ frames rather than individual frames. In this method, if there are $\alpha$ amount of fall frames, then the ground truth of the entire sequence of frames is labeled as a fall. The value of $\alpha$ is varied from $\alpha=1$ to $\alpha=T=8$. These anomaly scores are then used to compute the area under the curve  (AUC) of the ROC with fall as the class of interest.
The cross-context anomaly score gives a score per frame (for a given video); therefore, it can be directly compared with DAE and CAE variants. The anomaly scores obtained for every frame is used to calculate AUC of the ROC, with fall as the class of interest. For DAE and CAE variants, the reconstruction error of frames of a test video are considered as anomaly score.
The reported AUC is the average of AUC across all test videos. As discussed in Section \ref{sec:3DCAE-encdec}, we test three variants of DSTCAE: DSTCAE-UpSampling, DSTCAE-Deconv, DSTCAE-C3D.

\subsection{Results}
\label{sec:results}
\subsubsection{Cross-Context Anomaly Score}
The results for all frame-based models i.e. DAE, CAE variants, and DSTCAE variants using the cross context score, are presented in Table \ref{tab:all}. We also present the AUC values computed using an ensemble of one-class JK nearest neighbour method \cite{khan2018relationship} trained over the features learned from DAE; features from the fourth fully connected layer were used. The nearest neighbours are set to $1$ (JKNN-11) and $10$ (JKNN-JK) and the ensemble sizes is fixed to $25$. The results for UR and SDU using the one-class JK nearest neighbour method could not be obtained in reasonable time because it calculates euclidean distance for all pairs of input; thus making the method very slow and impractical to work with large datasets. We also compare the results from our previous work of using CLSTMAE  \cite{nogasfall2018} over all the datasets with cross-context scores $C_\sigma$ and $C_\mu$. The first column of Table \ref{tab:all} indicates the model tested, and the other columns show the dataset. Each numerical value in the Table \ref{tab:all} represents the average ROC AUC across all videos for the model indicated by the row. The respective standard deviation of ROC AUC across all videos of the dataset is shown in small brackets. 
The results can be summarized as follows

\begin{itemize}
    \item All the variants of spatio-temporal autoencoders outperforms spatial feature extraction methods (i.e. JKNN, DAE and two CAE variants). This confirms the fact that extracting both spatial and temporal features are important in video based fall detection approach.
    \item All the variants of DSTCAE outperforms the CLSTM based methods. This marks an improvement on our previous work \cite{nogasfall2018}. This may be the case since LSTM based models are tailored to long term sequential modelling, but falls occur on short time scales. It further confirms that DSTCAE is better suited for such problems of short length sequences.
    \item The autoencoders trained on holes filled versions of UR and SDU datasets performed better for all the classifiers (except for one CLSTMAE:$C_\sigma$ for the SDU dataset, which is worse than the best AUC obtained for this dataset with holes-filled). This result shows that if the depth data contains holes or missing pixels, it must be filled with some technique for the autoencoder to detect falls appropriately.
    \item The difference between DSTACE-Deconv and DSTCAE-UpSampling appears to be very small, but DSTCAE-C3D performs the best on two out of three datasets (excluding depth datasets without holes filled, which are degenerative cases).
    \item We see that $C_\sigma$ outperforms on $C_\mu$ on Thermal and UR-Filled data. That is, DSTCAE-C3D:$C_\sigma$, and DSTCAE-UpSampling:$C_\sigma$ perform the best on Thermal and UR-Filled data respectively.
\end{itemize}
We also perform a significance test to compare the ROC AUC of these different algorithms on all the 5 datasets (i.e. Thermal, UR, UR-filled, SDU and SDU-filled). In particular, we use the Friedman’s post-hoc test with Bergmann and Hommel’s correc- tion [4]. For simplicity, we choose to compare one variant from each type of algorithms (ie. DAE and one each from JKNN, DAE, CAE, CLSTM, and DSTCAE variants). That is, we compare JKNN-JK, DAE, CAE-Deconv, CLSTMAE:$C_\mu$, and DSTCAE-C3D:$C_\sigma$. The results of this significance test can be found in Table \ref{tab:SigTest}. The ith row and jth column compares the algorithm specified in the ith row of the table, to the algorithm specified in the jth column of the table. Significant differences are shown in bold. We see that the only algorithm with a significant difference from the others is DSTCAE-C3D:$C_\sigma$ . This supports the above findings, that spatio-temporal autoencoder (DSTCAE and variants) outperforms spatial feature extraction methods. 

It should be noted that these results may be misleading, firstly because we have a small sample size for conducting a significance test (only five datasets). We note that for the SDU, and SDU-Filled dataset, the best performing model (DSTCAE-C3D:$C_\mu$, 0.95(0.04) AUC) does not perform much better than the much simpler DAE model (0.94(0.05) AUC). The SDU dataset videos contains simple and organic activities; falls always happened from standing, besides having no furniture or background objects in the scene. We hypothesize that due to these reasons the DAE model may be able to learn global features to detect falls comparably to the spatio-temporal network. However, the activities in the Thermal and UR datasets were complex; falls happened in various poses (e.g. falling from chair, falling from sitting and falling from standing), and the scene involved different objects in the background (e.g. bed, chair). Besides that, in the Thermal dataset, due to a person entering the scene, the pixel intensity would change values due to change in the heat in the environment. The proposed DSTCAE methods worked well under these diverse conditions to detect unseen falls.

\begin{table}[!ht]
\centering
\begin{tabular}{|l|lllll|}
\hline
 & JKNN-JK & DAE & CAE-Deconv & CLSTMAE:$C_\mu$. & DSTCAE-C3D:$C_\mu$ \\
\hline
JKNN-JK & n/a & 0.143 & 0.287 & 0.075 & {\bf 0.003} \\
DAE & 0.143 & n/a & 1.000 & 1.000 & 0.431 \\
CAE-Deconv & 0.287 & 1.000 & n/a & 1.000 & 0.431 \\
CLSTMAE:$C_\mu$. & 0.075 & 1.000 & 1.000 & n/a & 0.543 \\
DSTCAE-C3D:$C_\mu$ & {\bf 0.003} & 0.431 & 0.431 & 0.543 & n/a \\
\hline
\end{tabular}

\caption{Friedman post-hoc test with Bergmann and Hommel’s correction}
\label{tab:SigTest}
\end{table}

\begin{table}[!ht]
\centering
\begin{tabular}{|c||c||c|c||c|c|}
\hline
\rowcolor{Gray} \textbf{Models} & \textbf{Thermal} & \textbf{UR} & \textbf{UR-Filled} & \textbf{SDU} & \textbf{SDU-Filled} \\ \hline\hline
JKNN-11 & 0.53 &  - & 0.5 & - & 0.72\\ \hline
JKNN-JK & 0.53 &  - & 0.5 & - & 0.86\\ \hline
DAE & 0.65(0.14)&	0.38(0.14)&	0.75(0.15)&	0.72(0.16)&	0.94(0.05) \\ \hline
CAE-UpSampling & 0.73(0.12)&	0.38(0.14)&	0.67(0.19)&	0.64(0.16)&	0.89(0.06)\\ \hline
CAE-Deconv & 0.75(0.17)&	0.38(0.11)&	0.76(0.21)&	0.66(0.16)&	0.92(0.07) \\ \hline
CLSTMAE: $C_{\sigma}$ & 0.64(0.16)&	0.49(0.09)&	0.67(0.20)&	0.66(0.10)&	0.56(0.11) \\ \hline
CLSTMAE: $C_{\mu}$ & 0.58(0.20)&	0.43(0.11)&	0.82(0.16)&	\textbf{0.77(0.11)}&	0.92(0.06) \\ \hline
DSTCAE-UpSampling:$C_{\sigma}$ & 0.96(0.03)&	0.69(0.18)&	\textbf{0.89(0.09)}&	0.74(0.12)&	0.90(0.06) \\ \hline 
DSTCAE-UpSampling:$C_{\mu}$ & 0.95(0.04) &	0.50(0.14)&	0.88(0.12)&	0.70(0.18)&	0.92(0.05) \\ \hline
DSTCAE-Deconv:$C_{\sigma}$ & 0.96(0.02)&	\textbf{0.72(0.17)}& 0.88(0.10)&	0.70(0.13)&	0.92(0.05) \\ \hline
DSTCAE-Deconv:$C_{\mu}$ & 0.94(0.04)&	0.48(0.14)&	0.85(0.11)&	0.69(0.18) & 0.92(0.05) \\ \hline
DSTCAE-C3D:$C_{\sigma}$ & \textbf{0.97(0.02)}&	 0.54(0.13)& 0.80(0.13)& 0.76(0.11)& 0.92(0.06) \\ \hline
DSTCAE-C3D:$C_{\mu}$ & 0.93(0.07)&	0.45(0.12) & 0.86(0.17&	0.71(0.18)&	\textbf{0.95(0.04)} \\ \hline

\end{tabular}
\caption{ROC AUC values for different methods, for each dataset, based on cross-context anomaly score.}
\label{tab:all}
\end{table}

\subsubsection{Within-Context Anomaly Score}

\begin{figure}[!ht]
    \centering
  \captionsetup{justification=centering}
  \captionsetup[subfigure]{width=100pt, justification=centering}%
  \begin{subfloat}[Thermal dataset.]
    \centering
    \includegraphics[width=0.8\textwidth]{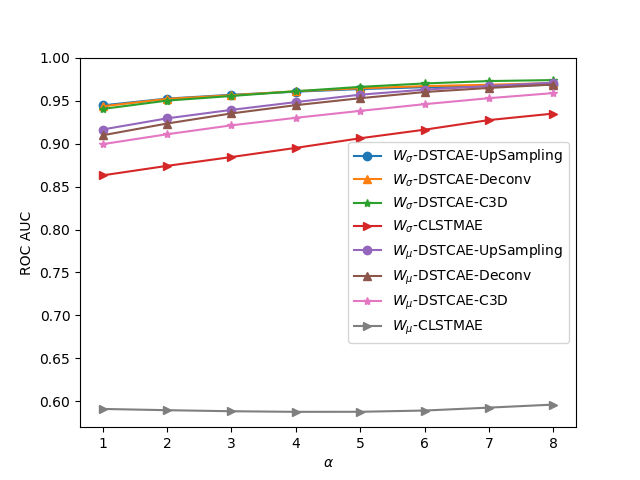}
    \label{fig:thermal}
  \end{subfloat}
  \begin{subfloat}[UR-Filled dataset.]
    \centering
    \includegraphics[width=0.8\textwidth]{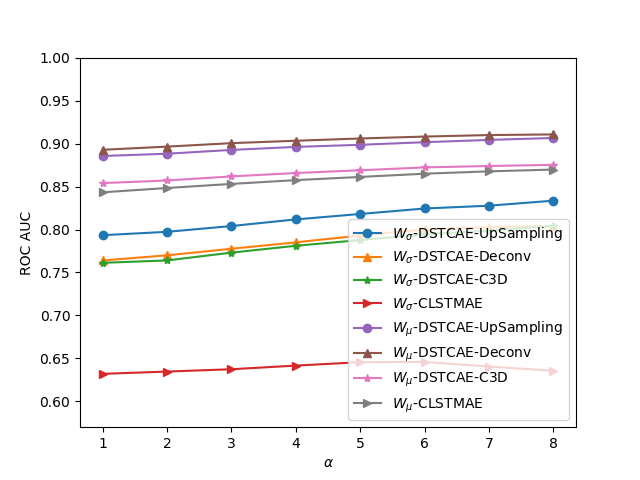}
    \label{fig:urhole}
  \end{subfloat}
  \begin{subfloat}[SDU-Filled dataset.]
    \centering
    \includegraphics[width=0.8\textwidth]{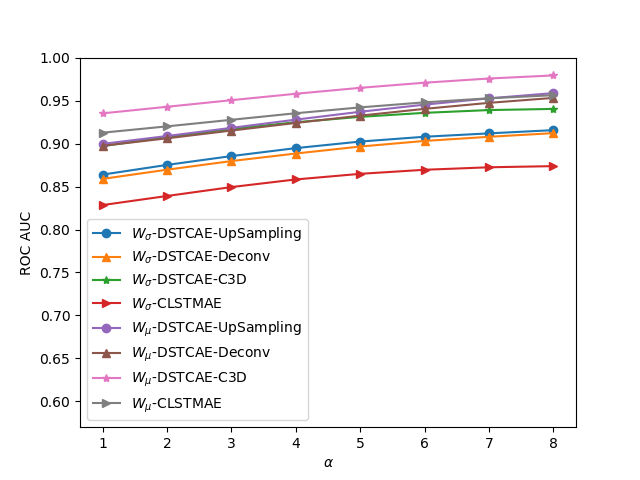}
    \label{fig:SDU-Filled_DSTCAE_tol}
  \end{subfloat}
  \centering
  \caption{Effect of changing fall frames in a stack of frames ($\alpha$) on the ROC AUC.}
  \label{fig:withincontext}
\end{figure}

The results achieved using the wintin-context anomaly scores, $W_\mu$ and $W_\sigma$, for all the datasets are shown in Figure \ref{fig:withincontext}. We do not show scores for SDU and UR with holes, because results were significantly worse than those with holes filled. We observe that increasing $\alpha$ increases the ROC AUC for Thermal and SDU-Filled datasets, and to a lesser extent, UR-Filled dataset. This indicates that reducing false positives is effective in increasing the ROC AUC score, and thus the ability to detect falls.

We also observe that the ROC AUC attained using $W_\mu$ and $W_\sigma$ (with sufficiently high $\alpha$) is at least as high, and in some cases higher than using frame based scores $C_\sigma$ and $C_\mu$. That is, for Thermal dataset, the highest ROC AUC remained consistent from within-context to cross-context scores. For UR-Filled, the ROC AUC improved from $0.89$ ($C_\sigma$) to $0.91$ ($W_\mu$), and for SDU-Filled dataset, ROC AUC improved from $0.95$ ($C_\sigma$) to $0.98$ ($W_\mu$). Scores for JKNN-11 and JKNN-JK are not shown in Figure \ref{fig:withincontext}, because these methods do not give scores on a per-frame basis, and do not give window based scores.

\section{Conclusions, Discussion, and Future Work}
\label{sec:conclusions}

Detecting falls in a non-invasive manner is a challenging problem; especially, as falls occur rarely. In this paper, we formulated detecting falls as an anomaly detection problem and presented the \textit{DeepFall} framework, which shows the development and comparison of three variants of DSTCAE. We tested the \textit{DeepFall} framework on three datasets that captured ADL and falls in a non-invasive manner using thermal and depth camera. We presented a new anomaly scoring method - referred to as the within-context anomaly score. The results showed that the \textit{DeepFall} framework outperforms the standard CAE, as well as traditional DAE. We also found that DSTCAE variants outperformed the convolutional LSTM autoencoder approch of detecting falls.. It was also observed that the within-context anomaly score, outperformed the per-frame based anomaly scoring method (or cross-context anomaly score) on two out of three datasets (excluding depth data sets without hole filling). 

The performance of the \textit{DeepFall} method may be affected by the presence of multiple persons in a scene, occurrence of other anomalous behaviour or lighting conditions. We are planning to collect a new dataset on fall detection that will address some of these issues to build a robust fall detection system

The main application of 3D Convolutional Spatio-temporal Autoencoders investigated in this paper is fall detection. A higher performance is achieved because the concept of normal activities of daily living was clearly defined. This idea can be further extended to other possible applications, such as anomalous behaviour in crowded scenes, violence detection in public places, detecting security threats in open areas. In all of these applications, individual facial or body movements are not required, rather generic spatio-temporal features can be learned from normal behaviour and anomalies may be flagged. 

Explainability of deep learning based methods is a challenge \cite{samek2017explainable}. The advantage with spatio-temporal convolutional methods is that the filters learned during the training process can be visualized. This will improve our understanding of the mechanism of fall detection. 
In the future, we will explore the application of Generative Adversarial Networks \cite{goodfellow2014generative} to the task of fall detection in an unsupervised setting. 

\section{Conflict of Interest}

On behalf of all authors, the corresponding author (Jacob Nogas) states that there is no conflict of interest.

\clearpage
\bibliographystyle{spmpsci}
\bibliography{references}

\end{document}